\documentclass[letterpaper]{article}

\ifdefined\aaaianonymous
    \usepackage[submission]{aaai2027}
\else
    \usepackage{aaai2027}
\fi

\usepackage{times}
\usepackage{helvet}
\usepackage{courier}
\usepackage[hyphens]{url}
\usepackage{graphicx}
\usepackage{natbib}
\usepackage{caption}
\usepackage{xcolor}
\usepackage{booktabs}
\usepackage{array}
\usepackage{multirow}
\usepackage{amsmath}
\usepackage{amssymb}
\usepackage{algpseudocode}
\usepackage{placeins}

\newcommand{\thead}[1]{\textbf{#1}}
\newcommand{\best}[1]{\textbf{#1}}
\newcommand{\ours}{SEAC}

\newcounter{algorithm}
\newenvironment{seacalgorithm}[2]{%
    \refstepcounter{algorithm}%
    \par\medskip\noindent\rule{\columnwidth}{0.4pt}\par
    \noindent\textbf{Algorithm~\thealgorithm: #1}\label{#2}\par\vspace{2pt}
}{%
    \par\noindent\rule{\columnwidth}{0.4pt}\par\medskip
}

\title{Stop or Clarify? Experience-Conditioned Active Fact Verification}
\author{
Yuhang Dong,
Jinghong Li,
Xinyue Wang,
Yuanning Cui\corresponding
}
\affiliations{Nanjing University of Information Science and Technology, Nanjing, China}

\begin{document}
\maketitle

\begin{abstract}
Fact-verification benchmarks score the final label, while an interactive verifier must first decide whether the available evidence is sufficient. We focus on that intermediate \emph{Stop-or-Clarify Decision}. Uncertainty alone is insufficient, because the benefit of another evidence-acquisition step is unknown before the step is taken. Sequential Experience-Adaptive Clarification (\ours{}) wraps a fixed verifier with a lightweight evidence-acquisition policy. For each claim, \ours{} retrieves evidence, runs the verifier, forms a label-free decision state, and combines stop risk with an observational clarification-utility estimate from earlier executed trajectories. Clarification is requested only when this score exceeds the action cost and decision threshold. On SciFact, LIAR, and FEVER, \ours{} improves mean accuracy and macro-F1 over the compared fact-verification and active-retrieval baselines. Equal-budget and formula-level diagnostics support selective clarification as a contributor to the gain, while the small SciFact split leaves its paired improvement inconclusive.
\end{abstract}

\section{Introduction}

Existing fact verification research is typically formulated as evidence-grounded label prediction. Given a claim and supporting documents, a verifier decides whether the claim is supported, refuted, or unverifiable~\citep{thorne2018fever,wadden2020scifact,wang2017liar,tang2024minicheck,dehaven2023bevers}. That formulation has produced strong verifiers and retrieval-augmented pipelines~\citep{lewis2020rag,tan2025corxfact,lin2025factaudit}. In an interactive setting, however, label prediction is preceded by another decision. Before issuing a verdict, the verifier must decide whether the evidence already found is sufficient or whether one more evidence-acquisition action is worth the cost.

We refer to this intermediate action as the \emph{Stop-or-Clarify Decision}. The verifier can stop and return the current label, or it can clarify by issuing a targeted evidence request and verifying again. The two mistakes have different costs. Stopping early may lock in a wrong label before decisive evidence appears. Clarifying too often can add topical but non-decisive passages while consuming the interaction budget. The question is under what circumstances another evidence-acquisition step is expected to be beneficial.

The decision is difficult for two reasons. First, there is no reliable stop label. Active retrieval and agentic verification methods have improved how systems ask for more evidence~\citep{yao2023react,jiang2023active,trivedi2023ircot,gao2023rarr,asai2024selfrag,huo2025microact,xie2025fire}, but they rarely specify when the system should stop asking. A verifier observes retrieved passages and confidence scores. It does not observe whether the decisive evidence has already been found. Confidence is useful but incomplete: some low-confidence cases cannot be repaired by retrieval, and some confident cases hide a missing time, entity, or scope constraint.

Second, clarification value must be estimated before the action is taken. A risky state does not necessarily lead to a useful next query. Decision-theoretic work treats such choices as value-of-information problems~\citep{howard1966information,horvitz1987resource,russell1991right,krause2005value}. Offline reinforcement learning, contextual bandits, and doubly robust estimators can estimate action values when logged rewards or action propensities are available~\citep{levine2020offline,langford2008epoch,li2010contextual,dudik2011doubly}. Active fact verification does not provide that feedback during inference. The system observes only the evidence path it follows, while the label that would have resulted from the other action remains counterfactual. Gold labels enter only after prediction, for evaluation. Accordingly, the estimator must be derived from observable trajectories.

We present Sequential Experience-Adaptive Clarification (\ours{}). \ours{} adds a stop-or-clarify layer around an evidence-grounded verifier. For each claim, the verifier retrieves evidence and produces a label distribution. The layer then builds a compact state from confidence, evidence coverage, conflict cues, and remaining budget. From this state, \ours{} estimates the risk of stopping. It also retrieves similar earlier state--action--outcome trajectories and uses their observable utility to estimate whether clarification has helped in comparable situations.

Together, the two estimates define a cost-sensitive stop-or-clarify boundary. \ours{} selects \textsc{Clarify} only when a targeted evidence action is estimated to be worth its cost. Experience is used for this specific decision, not as a general task memory~\citep{graphiti2026,mem02026,nous2026hermes}. Compared with uncertainty-triggered repeated retrieval, experience changes where the system draws the boundary between stopping and asking.

Experiments on SciFact, LIAR, and FEVER use the same verifier, retriever, and label interface for all methods~\citep{wadden2020scifact,wang2017liar,thorne2018fever}. The shared setup isolates the stop-or-clarify decision. \ours{} improves accuracy and macro-F1 over fact-verification, active-retrieval, and control baselines. The gains persist under equal action budgets, and offline diagnostics show that trajectory utility is associated with post-hoc label improvement.

\paragraph{Contributions.}
\begin{itemize}
    \item The Stop-or-Clarify Decision is formalized for active fact verification: before returning a label, a system must decide whether the current evidence is sufficient.
    \item A cost-sensitive decision boundary is introduced to combine label-free stop risk with estimated clarification value under a bounded interaction budget.
    \item Clarification utility is estimated from online trajectories with executed clarification outcomes through a label-free observational proxy; no gold-label feedback is used during inference.
    \item We provide shared-interface experiments and diagnostics covering final-label quality, equal-budget behavior, and the association between trajectory utility and post-hoc label changes.
\end{itemize}

\section{Related Work and Positioning}

\paragraph{Fact verification.}
Evidence-grounded fact verification judges a claim against retrieved or provided evidence~\citep{thorne2018fever,augenstein2019multifc,wadden2020scifact,jiang2020hover,schuster2021vitaminc,wang2017liar}. Work in this area has improved several parts of the pipeline: compact grounding verifiers reduce inference cost~\citep{tang2024minicheck}, general frameworks improve robustness across claim types~\citep{dehaven2023bevers}, and LLM-based systems add explanation, conflict reasoning, knowledge-graph checking, or auditability~\citep{tan2025corxfact,qudus2025kgfact,lin2025factaudit}. Prior methods mainly improve label prediction after evidence has been selected. \ours{} studies the preceding decision of whether that evidence should be accepted or expanded.

\paragraph{Active retrieval.}
Retrieval-augmented generation and dense retrieval established the value of external evidence for knowledge-intensive reasoning~\citep{lewis2020rag,karpukhin2020dpr}. Active learning gives a related principle: information should be acquired selectively when it carries a cost~\citep{settles2009active}. Active and agentic retrieval systems revise queries, interleave reasoning with retrieval, or critique generated content~\citep{yao2023react,jiang2023active,trivedi2023ircot,gao2023rarr,asai2024selfrag,huo2025microact,xie2025fire}. Such flexibility improves evidence acquisition, but it does not settle the stopping question. If uncertainty alone triggers another retrieval step, the system can spend budget on passages that are topical rather than decisive. \ours{} treats evidence acquisition as an action whose estimated value must exceed its cost.

\paragraph{Decision under cost.}
The stop-or-clarify decision follows the same resource logic as value-of-information and resource-bounded reasoning: acquiring information is useful only when expected utility exceeds cost~\citep{howard1966information,horvitz1987resource,russell1991right,krause2005value}. Reinforcement learning, offline RL, contextual bandits, and doubly robust estimators can learn action values when rewards or propensities are logged~\citep{sutton2018reinforcement,levine2020offline,langford2008epoch,li2010contextual,dudik2011doubly}. Active fact verification typically lacks that feedback during inference. \ours{} uses trajectory evidence as a label-free approximation, without claiming full counterfactual identification. Persistent memory systems show that experience can be retained across tasks~\citep{graphiti2026,mem02026,nous2026hermes}; here, experience is used more narrowly to estimate observable clarification utility for the current boundary.

\section{Method}

\subsection{Task Setup and Decision Interface}

\begin{figure*}[t]
    \centering
    \includegraphics[width=0.86\textwidth]{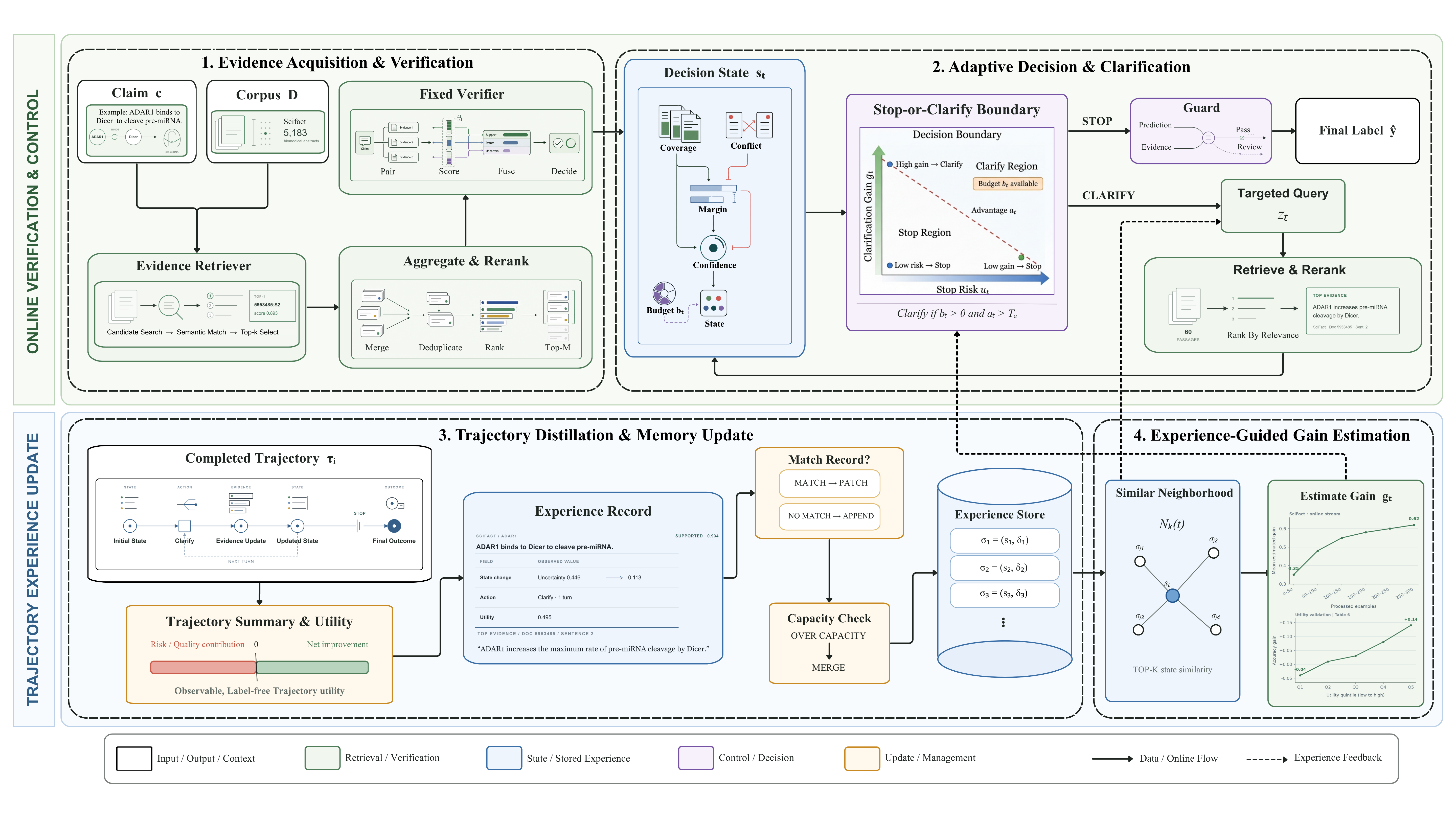}
    \caption{\ours{} framework overview. A fixed verifier predicts a label from the current evidence. The stop-or-clarify decision module estimates stop risk and clarification value, then either accepts the current label or executes one focused clarification step. Completed trajectories update later gain estimates.}
    \label{fig:seac-framework}
\end{figure*}

Figure~\ref{fig:seac-framework} shows the overall flow of \ours{} around \emph{active fact conflict verification}. The input is a claim \(c\) and an evidence corpus \(\mathcal{D}\). The output is a fact label in the shared label space \(\mathcal{Y}=\{\textsc{Supported},\textsc{Refuted},\textsc{NEI}\}\). Here \(\textsc{NEI}\) denotes that the available evidence is insufficient to support or refute the claim. Dataset-specific labels are converted into \(\mathcal{Y}\) only to create a common verifier interface. The exact mappings are reported in Appendix~\ref{app:dataset-adaptation}.

The intermediate decision is defined after an initial verifier prediction. At decision step \(t\), let \(\mathcal{E}_t\subset\mathcal{D}\) be the current evidence set and \(b_t\) the remaining clarification budget. The method chooses \(d_t\in\{\textsc{Stop},\textsc{Clarify}\}\). If \(d_t=\textsc{Stop}\), inference terminates and the current label is returned. If \(d_t=\textsc{Clarify}\), one unit of \(b_t\) is spent, focused evidence is retrieved, and the verifier is called again. The verifier predicts labels; \ours{} decides whether the current prediction is ready to be accepted.

\subsection{Evidence Retrieval and Base Verification}

\ours{} begins with a standard retrieval-and-verification pass. Each passage in \(\mathcal{D}\) is embedded offline, normalized to unit length, and stored in a FAISS inner-product index. For a query \(q\), \(\mathrm{RETRIEVE}(q,\mathcal{D})\) embeds \(q\) with the same encoder, normalizes the query vector, and returns the top-\(\mathrm{K}\) passages by inner product. The first query is the claim itself, giving \(\mathcal{E}_0=\mathrm{RETRIEVE}(c,\mathcal{D})\).

Given the current pair \((c,\mathcal{E}_t)\), the fixed verifier returns a label distribution \(\mathbf{p}_t\) and a current prediction \(\hat{y}_t\):
\begin{equation}
\mathbf{p}_t=\mathrm{VERIFY}(c,\mathcal{E}_t),\qquad
\hat{y}_t=\arg\max_{y\in\mathcal{Y}}\mathbf{p}_t(y).
\end{equation}
Here \(\mathbf{p}_t(y)\) is the verifier probability for label \(y\), and \(\hat{y}_t\) is the label returned if the method stops at decision step \(t\). The verifier itself is not trained or modified; its output becomes the input to the stop-or-clarify decision.

\subsection{Stop Risk from the Current State}

Evidence sufficiency is latent, so the stop signal cannot depend on the gold label. \ours{} builds a label-free decision state from quantities available at inference time. The state and stop-risk score are
\begin{equation}
\mathbf{s}_t=\mathrm{FEAT}(c,\mathcal{E}_t,\mathbf{p}_t,b_t),\qquad
u_t=\mathrm{RISK}(\mathbf{s}_t).
\end{equation}
The vector \(\mathbf{s}_t\) contains verifier confidence, prediction margin, neutral probability, evidence coverage, conflict cues, and remaining budget. The scalar \(u_t\in[0,1]\) estimates the risk of accepting \(\hat{y}_t\) immediately. Larger values indicate weak evidence, verifier uncertainty, or unresolved conflicts in the retrieved passages. The function \(\mathrm{RISK}\) is deterministic and label-free; its feature normalization and coefficients are fixed before final evaluation and are not updated from evaluation labels.

\subsection{Stop-or-Clarify Boundary}

Stop risk alone cannot decide the action, because some risky states remain unrecoverable after another retrieval step. Clarification utility is estimated separately. Let \(g_t\in\mathbb{R}\) denote the expected observable utility of clarification at state \(\mathbf{s}_t\), estimated from previous executed clarifications as described below. Let \(\mathrm{C}_{\mathrm{ask}}\) be the fixed cost of one clarification turn. \ours{} uses the following clarification score:
\begin{equation}
a_t=u_t+g_t-\mathrm{C}_{\mathrm{ask}}.
\end{equation}
The scalar \(a_t\) is a one-step decision score for asking versus stopping. It follows a resource-bounded information-acquisition view: stop risk approximates the loss of accepting the current label, trajectory evidence summarizes observable improvement after similar clarifications, and the cost term prevents extra turns from being treated as free~\citep{howard1966information,horvitz1987resource,russell1991right,krause2005value}. Because the trajectory term is observational, \(a_t\) is neither an identified treatment effect nor an exact value-of-information quantity. More expressive estimators, including reinforcement learning, offline RL, contextual bandits, doubly robust estimation, or neural policies, can be used when logged rewards, action propensities, or counterfactual supervision are available~\citep{sutton2018reinforcement,levine2020offline,langford2008epoch,li2010contextual,dudik2011doubly,chen2021decision}. We instead use a label-free proxy and evaluate its association with label changes only after inference.

The action \(d_t\) is then chosen by a cost-sensitive boundary:
\begin{equation}
d_t=
\begin{cases}
\textsc{Clarify}, & b_t>0 \ \text{and}\ a_t>\mathrm{T}_{\mathrm{A}},\\
\textsc{Stop}, & \text{otherwise},
\end{cases}
\end{equation}
where \(\mathrm{T}_{\mathrm{A}}\) is a fixed threshold constant. The corresponding boundary is
\begin{equation}
\mathcal{B}_t=\{\mathbf{s}: u(\mathbf{s})+g(\mathbf{s})-\mathrm{C}_{\mathrm{ask}}=\mathrm{T}_{\mathrm{A}}\}.
\end{equation}
The boundary \(\mathcal{B}_t\) is the object the method adapts. Two states may have similar verifier uncertainty but different clarification value. When past trajectories show that clarification helps one kind of ambiguity but not another, \(g_t\) places those states on different sides of \(\mathcal{B}_t\).

\subsection{Clarification and Trajectory Evidence}

When the boundary selects \(d_t=\textsc{Clarify}\), \ours{} performs one focused evidence-acquisition step. The query routine \(z_t=\mathrm{QUERY}(c,\mathcal{E}_t,\mathcal{N}_{\mathrm{K}}(t))\) receives the claim \(c\), the current evidence \(\mathcal{E}_t\), and a local neighborhood \(\mathcal{N}_{\mathrm{K}}(t)\) of similar past trajectories. It identifies one unresolved ambiguity and rewrites it as a short retrieval query \(z_t\). The refreshed evidence is
\begin{equation}
\mathcal{E}_{t+1}=\mathrm{RERANK}(\mathcal{E}_t\cup\mathrm{RETRIEVE}(z_t,\mathcal{D}),c).
\end{equation}
\(\mathrm{RERANK}\) removes duplicates, scores candidate passages by claim relevance and verifier compatibility, and keeps a bounded evidence set for the next verifier call. The verifier is then called again on \((c,\mathcal{E}_{t+1})\). Prompt templates and guard paths are implementation details reported in the artifact and Appendix; the method here is the action decision.

Only executed clarifications provide an observed post-action state. The gain memory therefore excludes stop decisions, whose clarification outcomes are unobserved. A stored record is \(\sigma_i=(\mathbf{s}_i,\textsc{Clarify},\delta_i)\), where \(\mathbf{s}_i\) is the pre-action state and \(\delta_i\) is a label-free utility proxy computed after clarification. Let \(u_i^0\) and \(u_i^1\) denote stop risk before and after the action. Let \(q_i^0\) and \(q_i^1\) denote deterministic evidence-quality summaries in \([0,1]\), computed from normalized retrieval relevance, claim coverage, and cross-passage consistency. Let \(n_i\) be the number of clarification turns represented by the record. The stored utility is
\begin{equation}
\delta_i=(u_i^0-u_i^1)+(q_i^1-q_i^0)-\mathrm{C}_{\mathrm{ask}}n_i .
\end{equation}
The utility does not use the gold label. It records whether the observable state improved enough to justify the cost. Stop actions remain in the audit trace but do not enter the gain estimator. Labels are used later only for offline analysis, where we test whether higher \(\delta_i\) is associated with wrong-to-correct transitions.

Before deciding at state \(\mathbf{s}_t\), \ours{} retrieves the \(\mathrm{K}\) most similar previous records:
\begin{equation}
\begin{aligned}
r_i(t)&=\mathrm{SIM}(\mathbf{s}_t,\mathbf{s}_i),\\
\mathcal{N}_{\mathrm{K}}(t)&=\mathrm{TOPK}_{\mathrm{K}}\{r_i(t):\sigma_i\in\mathcal{M}_t\}.
\end{aligned}
\end{equation}
Here \(r_i(t)\) is the similarity between the current state and record \(i\). The gain estimate is
\begin{equation}
g_t=\frac{\sum_{\sigma_i\in\mathcal{N}_{\mathrm{K}}(t)} r_i(t)\delta_i}
{\mathrm{EPS}+\sum_{\sigma_i\in\mathcal{N}_{\mathrm{K}}(t)} r_i(t)} ,
\end{equation}
where \(\mathrm{EPS}\) prevents division by zero. This is a local Nadaraya--Watson estimator of \(\mathbb{E}[\delta\mid\mathbf{s}_t,d=\textsc{Clarify}]\), with state similarity acting as a kernel~\citep{cover1967nearest,nadaraya1964estimating,watson1964smooth}. The estimator relies on overlap and local smoothness: the memory must contain sufficiently similar clarified states, and similar verifier--evidence states must have similar observable outcomes~\citep{stone1977consistent}. It does not estimate clarification utility in regions unsupported by executed actions. Shuffled-experience and static-gain controls test whether local matching contributes beyond a global prior.

\subsection{Final Prediction and Trace}

Inference ends when the method stops or exhausts the budget. Let \(\tau\) denote this terminal turn:
\begin{equation}
\tau=\min\{t: d_t=\textsc{Stop}\ \text{or}\ b_t=0\},\qquad
\hat{y}=\hat{y}_{\tau}.
\end{equation}
The output is the final label \(\hat{y}\) and the decision trace \(\{d_t,a_t,u_t,g_t,b_t\}_{t\leq\tau}\). The trace records how the label was produced and supports the diagnostics in Section~\ref{sec:results}.

Algorithm~\ref{alg:compact-step} summarizes the inference step used by the boundary. It is placed here because the action is determined by the same quantities defined above: the current state \(\mathbf{s}_t\), the remaining budget \(b_t\), the risk score \(u_t\), and the trajectory-based gain \(g_t\). Evidence is updated only after the boundary selects \textsc{Clarify}; otherwise, the current verifier label is returned. The full online loop is given in Appendix~\ref{app:algorithm}.

\begin{seacalgorithm}{Compact Stop-or-Clarify Step for the boundary decision}{alg:compact-step}
{\footnotesize
\begin{algorithmic}[1]
\Require claim \(c\), evidence \(\mathcal{E}_t\), state \(\mathbf{s}_t\), budget \(b_t\), memory \(\mathcal{M}_t\)
\State \(u_t\gets\mathrm{RISK}(\mathbf{s}_t)\)
\State \(g_t\gets\mathrm{GAIN}(\mathrm{TOPK}_{\mathrm{K}}(\mathcal{M}_t,\mathbf{s}_t))\)
\State \(a_t\gets u_t+g_t-\mathrm{C}_{\mathrm{ask}}\)
\If{\(b_t>0\) and \(a_t>\mathrm{T}_{\mathrm{A}}\)}
    \State \(z_t\gets\mathrm{QUERY}(c,\mathcal{E}_t,\mathcal{M}_t)\)
    \State \Return \(\textsc{Clarify}\) with refreshed evidence
\EndIf
\State \Return \(\textsc{Stop}\) with current verifier label
\end{algorithmic}
}
\end{seacalgorithm}

\section{Experiments}

\subsection{Datasets}

Experiments use three fact-verification datasets. SciFact contains scientific claims grounded in biomedical abstracts~\citep{wadden2020scifact}. LIAR contains short political statements collected from PolitiFact~\citep{wang2017liar}. FEVER contains Wikipedia-based claims with evidence annotations~\citep{thorne2018fever}. Because the datasets were built for related but different verification settings, we expose their labels and evidence fields through one interface before comparing stop-or-clarify policies.

Each record is converted into \(\langle c,y,\mathcal{G},m\rangle\), where \(c\) is the claim, \(y\in\mathcal{Y}\) is the mapped label, \(\mathcal{G}\) is the available gold evidence used only for analysis, and \(m\) stores dataset metadata. The shared label space is \(\mathcal{Y}=\{\textsc{Supported},\textsc{Refuted},\textsc{NEI}\}\). The main evaluation includes 300 SciFact development claims with a 5,183-abstract corpus, 2,297 LIAR validation examples after adapter filtering, and 19,998 FEVER development examples. The adapted class counts are 124/64/112 \textsc{Supported}/\textsc{Refuted}/\textsc{NEI} examples for SciFact, 601/1,325/371 for LIAR, and 6,666/6,666/6,666 for FEVER. Exact mappings are reported in Appendix~\ref{app:dataset-adaptation}.

\subsection{Compared Methods}

To isolate the stop-or-clarify decision, every compared method operates on the same corpus, label space, and interaction budget. Official scores from the original papers are not directly comparable here, since those papers use different task interfaces. We rerun or adapt each method in a shared harness. Methods that require retrieval use the same FAISS index and evidence corpus. Methods that require verification use the same three-way label interface.

We organize baselines into three groups. Control baselines measure simple alternatives: Random maps a deterministic hash of the claim to one of the three labels, Retrieval-only predicts from the shared top-\(\mathrm{K}\) evidence using retrieval score and lexical conflict cues, and Direct LLM prompts the local LLM with the claim but no retrieved evidence. Fact-verification baselines test label prediction without an adaptive stop-or-clarify boundary: MiniCheck~\citep{tang2024minicheck}, BEVERS~\citep{dehaven2023bevers}, CorXFact~\citep{tan2025corxfact}, and FACT-AUDIT~\citep{lin2025factaudit}. Active baselines test whether iterative evidence acquisition is sufficient: Active MiniCheck uses uncertainty-triggered clarification, Micro-Act~\citep{huo2025microact} and FIRE~\citep{xie2025fire} use their active retrieval strategies, ReAct-Agent~\citep{yao2023react} uses LLM-guided tool calls, and Safe-Agent is a conservative local adapter that suppresses low-confidence tool actions.

The harness is designed for a controlled decision comparison, not for reproducing every original leaderboard setting. Appendix~\ref{app:dataset-adaptation} gives the label mappings, prompt adapters, and baseline-specific details needed for reproduction.

\subsection{Evaluation Protocol}

Each run is defined by a dataset, a method, and a seed. For a run, the retrieval index and verifier are fixed before inference begins. The experience store starts empty, and examples are processed in split order. When making a decision for an example, a method can access the claim, retrieved evidence, verifier scores, remaining budget, and trajectories produced by earlier examples in the same stream. Gold labels are read only after the final prediction is returned.

Clarification budgets are part of the inference condition. The main setting fixes the maximum number of clarification turns at two for SciFact and one for LIAR and FEVER before metrics are computed. A method may stop earlier than the cap. Passive baselines have no clarification action. In the equal-budget comparison, every active baseline receives the same maximum cap as \ours{}. The SciFact budget curve is a diagnostic experiment that varies the cap after the main setting is fixed.

Main results aggregate five fixed seeds from 42 to 46. Seeds affect stochastic baselines, cache tie-breaking, and generation diagnostics, not the data split. Accuracy is the primary metric because all datasets are evaluated as three-way classification. Macro-F1 is computed per seed and then averaged. Efficiency is measured by average clarification turns and early-stop rate under the dataset budget cap.

The protocol prevents label leakage into the stop-or-clarify boundary. A gain-memory record contains \(\mathbf{s}_i\), the executed \textsc{Clarify} action, and \(\delta_i\), not the correct label; stop decisions are retained only in the audit trace. If clarification improves an observable verifier state but still leads to a wrong final label, the stored utility records only the observable state change. Pairwise diagnostics use exact McNemar tests~\citep{mcnemar1947note}, paired comparisons, and 2,000-sample bootstrap confidence intervals~\citep{efron1993bootstrap} with \(\alpha=0.05\), followed by Holm correction across datasets.

\subsection{Implementation}

The reported system uses MiniCheck RoBERTa-Large as the grounding verifier~\citep{tang2024minicheck}, BGE-small-en-v1.5 dense retrieval~\citep{xiao2023cpack}, FAISS inner-product search over normalized 384-dimensional vectors~\citep{johnson2017faiss}, and local Qwen2.5-7B-Instruct for arbitration and clarification generation~\citep{yang2024qwen25}\footnote{\url{https://huggingface.co/Qwen/Qwen2.5-7B-Instruct}}. Retrieval uses top-\(\mathrm{K}=10\) initially and refresh top-\(\mathrm{K}=10\) after clarification; trajectory estimation uses 20 neighbors, \(\mathrm{C}_{\mathrm{ask}}=0.10\), and \(\mathrm{T}_{\mathrm{A}}=0.56\). Qwen generation uses at most 128 new tokens with temperature 0.2. Experiments run in a PyTorch environment on two NVIDIA 4090D GPUs.

\section{Results}
\label{sec:results}

\begin{table*}[!t]
\centering
\caption{Evaluation-set accuracy/macro-F1 comparison. Higher is better. This table reports final-label quality under the shared verifier and retrieval interface.}
\label{tab:main-results}
\scriptsize
\setlength{\tabcolsep}{4.8pt}
\renewcommand{\arraystretch}{1.02}
\begin{tabular}{@{}llccc@{}}
\toprule
\thead{Group} & \thead{Method} & \thead{SciFact} & \thead{LIAR} & \thead{FEVER} \\
& & \multicolumn{3}{c}{\scriptsize Accuracy / Macro-F1} \\
\midrule
\multirow{3}{*}{Control}
& Random & .310 / .295 & .323 / .301 & .333 / .333 \\
& Retrieval-only & .400 / .288 & .517 / .356 & .332 / .242 \\
& Direct LLM & .453 / .439 & .363 / .341 & .556 / .544 \\
\addlinespace[2pt]
\multirow{4}{*}{Fact verification}
& MiniCheck & .363 / .310 & .592 / .481 & .553 / .489 \\
& BEVERS & .313 / .262 & .602 / .481 & .508 / .443 \\
& CorXFact & .657 / .627 & .633 / .473 & .609 / .554 \\
& FACT-AUDIT & .667 / .650 & .456 / .412 & .604 / .602 \\
\addlinespace[2pt]
\multirow{5}{*}{Active clarification}
& Active MiniCheck & .370 / .318 & .592 / .481 & .414 / .336 \\
& Micro-Act & .367 / .313 & .592 / .481 & .397 / .328 \\
& FIRE & .370 / .317 & .592 / .481 & .415 / .335 \\
& ReAct-Agent & .373 / .320 & .592 / .481 & .416 / .339 \\
& Safe-Agent & .373 / .320 & .592 / .481 & .416 / .338 \\
\midrule
Ours & Experience-conditioned SEAC
& \best{.713 / .702} & \best{.654 / .527} & \best{.640 / .635} \\
\bottomrule
\end{tabular}
\end{table*}

\paragraph{Main comparison.}
The main comparison tests whether a stop-or-clarify boundary improves the final fact label when evidence access is fixed. In Table~\ref{tab:main-results}, \ours{} has the highest mean accuracy on all three datasets, with gains of 0.046, 0.021, and 0.031 over the strongest compared baseline on SciFact, LIAR, and FEVER. The gains are not uniform. Clarification helps most when an added passage can resolve a missing condition in the claim, and less when evidence sparsity or verifier saturation leaves little room for another action to help.

The active baselines make this point clearer. Several of them retrieve or reason for additional turns, yet still trail the stronger passive verifiers. Extra evidence is not enough by itself; the question is whether the next evidence action is likely to change the verifier's judgment. Across seed runs, accuracy standard deviations remain below 0.007, so the trend is not driven by a single unstable run.

\begin{table*}[!t]
\centering
\caption{Paired significance against the strongest non-oracle baseline. The table tests whether the main comparison reflects reliable paired prediction changes.}
\label{tab:significance}
\includegraphics[width=0.86\textwidth]{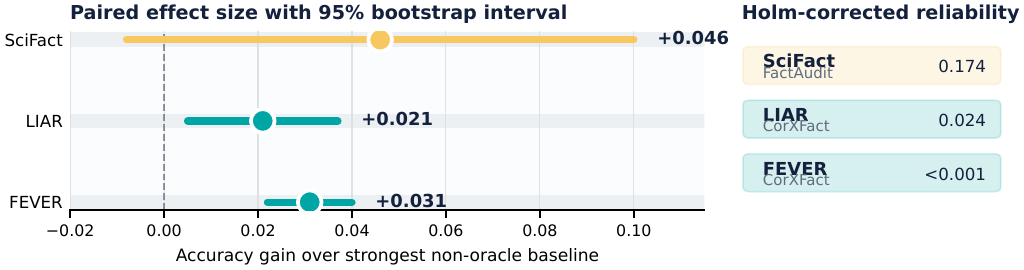}
\end{table*}

Significance is reported separately because it asks a paired question that the main table cannot answer: whether the prediction changes are reliable against the strongest compared baseline. Table~\ref{tab:significance} gives the clearest evidence on FEVER and LIAR, with Holm-corrected \(p\)-values of \(<0.001\) and 0.024. SciFact is positive but not conclusive, with a corrected \(p\)-value of 0.174; we therefore treat its effect size as directional rather than statistically established.

The table also separates evidence access from action value. Retrieval-only and active baselines can expose additional documents, but they do not estimate whether another clarification action is worth its cost. \ours{} filters evidence acquisition through a boundary, which is most useful when extra passages would add topical context without changing the verifier's judgment.

\begin{table}[t]
\centering
\caption{Equal-budget active comparison. \(B\) is the shared cap; Turn reports \ours{} / best active. This table separates action allowance from action selection.}
\label{tab:equal-budget-efficiency}
\scriptsize
\setlength{\tabcolsep}{3.0pt}
\renewcommand{\arraystretch}{1.04}
\begin{tabular}{@{}lcccc@{}}
\toprule
\thead{Data} & \thead{\(B\)} & \thead{\ours{} Acc.} & \thead{Best Active} & \thead{Turn} \\
\midrule
SciFact & 2 & \best{.713} & .390 & 1.00 / .28 \\
LIAR & 1 & \best{.654} & .592 & 1.00 / .50 \\
FEVER & 1 & \best{.640} & .416 & .64 / .60 \\
\bottomrule
\end{tabular}
\end{table}

\paragraph{Equal-budget comparison.}
Because \ours{} can request clarification, its main-table advantage could be misread as a larger action allowance. The equal-budget experiment addresses this concern. Every active baseline receives the same dataset-level clarification cap as \ours{} while retaining its own stopping rule and query strategy. The results are reported in Table~\ref{tab:equal-budget-efficiency}.

Giving every active baseline the same opportunity to ask does not close the gap. The gap is smallest on LIAR at 0.062 and largest on SciFact at 0.323. FEVER is especially informative: ReAct-Agent spends 0.60 turns on average, close to \ours{} in interaction count, yet its final labels remain weak. Because query generation also differs across methods, this experiment isolates the action cap rather than the boundary alone: it rules out a larger nominal budget as the explanation, while the formula-level ablations below more directly test the boundary terms.

\begin{figure*}[t]
    \centering
    \includegraphics[width=0.92\textwidth]{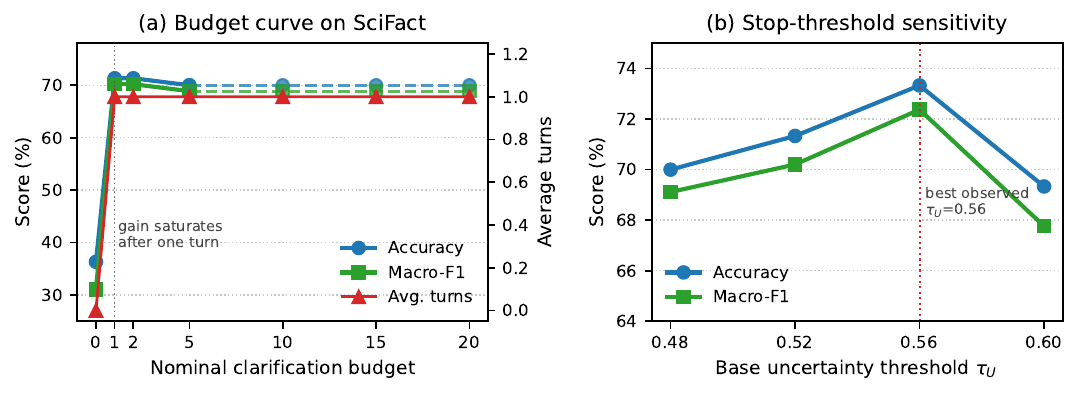}
    \caption{Focused SciFact diagnostics. Subfigure (a): clarification budget reaches early saturation rather than steady gains from repeated retrieval. Dashed budget points mark saturated reuse runs. Subfigure (b): threshold sensitivity exposes the role of the stop-or-clarify boundary; the best observed SciFact accuracy is reached without increasing average turn use.}
    \label{fig:scifact-diagnostics}
\end{figure*}

\paragraph{Budget and threshold behavior.}
Figure~\ref{fig:scifact-diagnostics} examines the boundary directly. Subfigure (a) varies the nominal clarification budget on SciFact. Accuracy moves from 0.363 at zero budget to 0.713 once clarification is available, then saturates because larger caps do not force extra interaction. A selective policy should behave this way: it spends a turn when useful, but does not turn a larger cap into repeated retrieval. Subfigure (b) changes the stop threshold while holding the verifier, retriever, and corpus fixed. The peak reaches 0.733, and accuracy falls when the boundary moves too far, supporting action selection as one component of the observed performance.

\begin{center}
\captionsetup{type=table}
\centering
\captionof{table}{Formula-level boundary ablations. The extended view reports accuracy change together with accuracy, turn, and stop-rate diagnostics.}
\label{tab:formula-ablations}
\includegraphics[width=0.99\columnwidth]{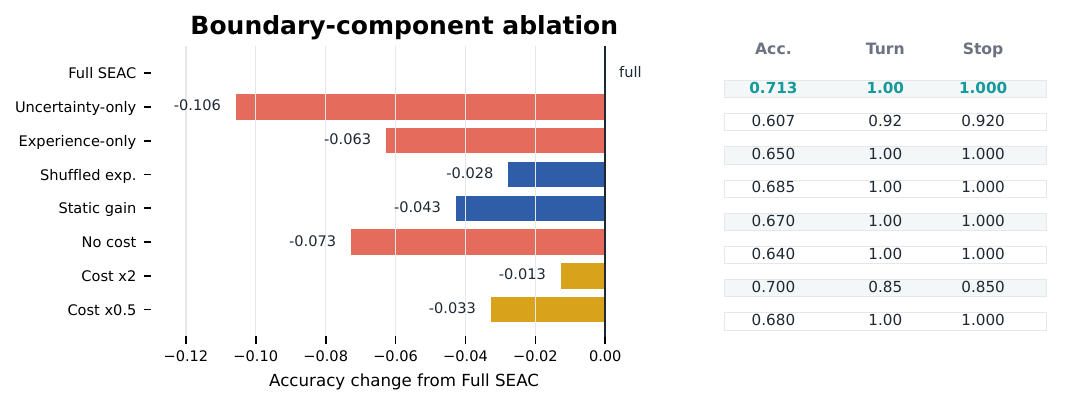}
\end{center}

\paragraph{Boundary component analysis.}
Formula-level ablations isolate the terms in the boundary rather than broad implementation switches. In Table~\ref{tab:formula-ablations}, neither stop risk nor trajectory evidence alone recovers the full result. The uncertainty-only variant reaches 0.607 on SciFact, while the experience-only variant reaches 0.650. The full boundary is stronger because it combines the current verifier state with matched trajectory evidence.

The matched-experience controls show what the trajectory term contributes. Shuffling trajectories lowers accuracy to 0.685, and replacing local gain with a static global mean lowers it further. The cost term gives the boundary its selective behavior: removing action cost produces the weakest SciFact ablation, consistent with over-clarification. The useful regime is neither ``always ask'' nor ``always stop'', but the middle region where stop risk, trajectory gain, and action cost jointly determine whether another evidence step is worthwhile.

These ablations also explain why coarse module switches are not enough for this paper's claim. A low-turn variant can still change labels through guard or arbitration paths, so the cleaner test is to alter the quantities that enter the boundary itself. The comparison therefore asks which decision evidence survives when the surrounding verifier, retriever, and generation interface remain fixed.

The comparison tables should be read together. Table~\ref{tab:main-results} reports final quality, Table~\ref{tab:significance} checks paired reliability, and Table~\ref{tab:equal-budget-efficiency} fixes the action unit. The strongest baseline differs across datasets, yet the same stop-or-clarify interface remains competitive. More retrieval-oriented behavior is therefore not automatically better; the useful comparison is accuracy under a visible action trace. Appendix~\ref{app:additional-diagnostics} extends this trace-level view with utility bins, transition counts, and failure decomposition.

Read together, Figure~\ref{fig:scifact-diagnostics} and Table~\ref{tab:formula-ablations} give the most direct evidence for the boundary view. The budget curve shows that extra allowance is useful only at the point where the policy first gains access to clarification. After that point, larger caps do not translate into repeated asking. The ablations explain why this happens: the cost term prevents clarification from becoming the default action, while the local trajectory term helps distinguish states where another query has historically changed the verifier state from states where it has not.

This pairing is important for interpreting the positive main results. A method that only raises the budget can improve when the missing evidence is easy to retrieve, but it has no reason to stop when the next passage is merely topical. Conversely, a method that only tightens the stop threshold can avoid cost, but may miss cases where a small evidence change repairs the label. SEAC is designed for the region between these two behaviors, where the trace records whether the gain came from acquiring new evidence or from refusing an unnecessary action.

\section{Conclusion}

We studied active fact verification through the Stop-or-Clarify Decision. \ours{} separates this action from label prediction and combines stop risk, local outcomes of executed clarifications, and explicit action cost. Under a shared evaluation interface, it obtains the highest mean accuracy among the compared methods; paired evidence is significant on LIAR and FEVER, while SciFact remains directional because of its small split. The trajectory term is observational and coverage-dependent, not a counterfactual value estimate.

Active-verification benchmarks should therefore report action traces together with final labels. Evidence sets, remaining budgets, executed actions, and pre/post verifier states would reveal whether gains come from decisive retrieval or from avoiding unnecessary interaction. Such logs would also support stronger bandit or supervised action-value estimators without changing the underlying fact-label task.

\FloatBarrier
\clearpage
\bibliography{references}

\clearpage
\appendix

\section{Online Inference Algorithm}
\label{app:algorithm}

\begin{seacalgorithm}{Online Stop-or-Clarify Inference}{alg:online-inference}
\begin{algorithmic}[1]
\Require claim \(c\), corpus \(\mathcal{D}\), store \(\mathcal{M}_0\), budget \(b_0\)
\Ensure final label \(\hat{y}\), completed trace \(\tau\), and updated store \(\mathcal{M}\)
\State \(\mathcal{E}_0\gets\mathrm{RETRIEVE}(c,\mathcal{D})\); \(\tau\gets\emptyset\); \(t\gets0\)
\State \(\mathbf{p}_0\gets\mathrm{VERIFY}(c,\mathcal{E}_0)\); \(\hat{y}_0\gets\arg\max_{y\in\mathcal{Y}}\mathbf{p}_0(y)\)
\State \(\mathbf{s}_0\gets\mathrm{FEAT}(c,\mathcal{E}_0,\mathbf{p}_0,b_0)\); \(u_0\gets\mathrm{RISK}(\mathbf{s}_0)\); \(q_0\gets\mathrm{QUALITY}(c,\mathcal{E}_0,\mathbf{p}_0)\)
\While{true}
    \State \(\mathcal{N}_{\mathrm{K}}(t)\gets\mathrm{TOPK}_{\mathrm{K}}(\mathcal{M}_t,\mathbf{s}_t)\); \(g_t\gets\mathrm{GAIN}(\mathcal{N}_{\mathrm{K}}(t),\mathbf{s}_t)\)
    \State \(a_t\gets u_t+g_t-\mathrm{C}_{\mathrm{ask}}\)
    \If{\(b_t=0\) \textbf{or} \(a_t\le \mathrm{T}_{\mathrm{A}}\)}
        \State Append \((\mathbf{s}_t,\textsc{Stop},a_t)\) to \(\tau\); \Return \(\hat{y}_t,\tau,\mathcal{M}_t\)
    \EndIf
    \State \(z_t\gets\mathrm{QUERY}(c,\mathcal{E}_t,\mathcal{N}_{\mathrm{K}}(t))\)
    \State \(\mathcal{E}_{t+1}\gets\mathrm{RERANK}(\mathcal{E}_t\cup\mathrm{RETRIEVE}(z_t,\mathcal{D}),c)\)
    \State \(b_{t+1}\gets b_t-1\); \(\mathbf{p}_{t+1}\gets\mathrm{VERIFY}(c,\mathcal{E}_{t+1})\); \(\hat{y}_{t+1}\gets\arg\max_y\mathbf{p}_{t+1}(y)\)
    \State \(\mathbf{s}_{t+1}\gets\mathrm{FEAT}(c,\mathcal{E}_{t+1},\mathbf{p}_{t+1},b_{t+1})\); \(u_{t+1}\gets\mathrm{RISK}(\mathbf{s}_{t+1})\)
    \State \(q_{t+1}\gets\mathrm{QUALITY}(c,\mathcal{E}_{t+1},\mathbf{p}_{t+1})\); \(\delta_t\gets(u_t-u_{t+1})+(q_{t+1}-q_t)-\mathrm{C}_{\mathrm{ask}}\)
    \State \(\mathcal{M}_{t+1}\gets\mathcal{M}_t\cup\{(\mathbf{s}_t,\textsc{Clarify},\delta_t)\}\)
    \State Append \((\mathbf{s}_t,\textsc{Clarify},a_t,\delta_t)\) to \(\tau\); \(t\gets t+1\)
\EndWhile
\end{algorithmic}
\end{seacalgorithm}

\section{Dataset Adaptation and Diagnostics}
\label{app:dataset-adaptation}

\paragraph{Shared label space.}
The shared interface uses the label space \(\{\textsc{Supported},\textsc{Refuted},\textsc{NEI}\}\). SciFact maps \textsc{Support} to \textsc{Supported}, \textsc{Contradict} to \textsc{Refuted}, and claims without annotated supporting or contradicting evidence to \textsc{NEI}. LIAR is adapted through the LIAR2 interface: labels 5/4 map to \textsc{Supported}, label 3 maps to \textsc{NEI}, and labels 2/1/0 map to \textsc{Refuted}. FEVER maps the official \textsc{Supports}, \textsc{Refutes}, and \textsc{NotEnoughInfo} labels directly to the three-way space.

\paragraph{Split usage.}
SciFact is evaluated on its 300-claim development split. The local resource also contains 809 training claims, 300 test claims, and a 5,183-abstract evidence corpus. LIAR uses 2,297 validation examples after adapter filtering. FEVER uses the 19,998-example shared-task development split. Experience stores are reset per dataset and seed. During evaluation, records can be created only from previous inference trajectories in the same stream. The current gold label, future examples, and evaluation labels are never used for stop-or-clarify decisions.

\paragraph{Baseline adaptation.}
MiniCheck and BEVERS are run from local copies of their released implementations. Micro-Act, FIRE, and ReAct-Agent use local adapters built from their public prompt and tool interfaces; Safe-Agent is our conservative tool-use control. Whenever a baseline permits it, the adapter exposes the same retrieval corpus and verifier interface used by \ours{}. CorXFact and FACT-AUDIT are reproduced as task-level adapters under the shared three-way label space, rather than copied from their original tables. The similar LIAR scores for active baselines are a property of this shared setup: short political claims and sparse evidence often leave the verifier unchanged after the first retrieval. We do not claim that the original methods are equivalent.

\paragraph{Budget diagnostics.}
Budget is an inference-time cap fixed before final evaluation. The focused SciFact budget curve executes or reuses budgets 0, 1, 2, and 5. Budgets 10, 15, and 20 reuse the saturated budget-5 run because that run consumed exactly one turn for every example. They are included only to show that larger nominal caps do not force additional clarification.

\section{Additional Diagnostics}
\label{app:additional-diagnostics}

\paragraph{Ablation-path audit.}
Table~\ref{tab:ablation-audit} records why the earlier coarse ablations are treated as diagnostics rather than causal evidence. No-clarification and no-graph variants can still change labels through guard or arbitration paths, so Turn alone does not explain those rows. The audit motivates the formula-level ablations reported in the main text.

\begin{table}[!htbp]
\centering
\caption{Ablation-path audit. Percentages report how often non-clarification paths changed or supported the final decision.}
\label{tab:ablation-audit}
\scriptsize
\setlength{\tabcolsep}{3pt}
\renewcommand{\arraystretch}{1.02}
\resizebox{\columnwidth}{!}{%
\begin{tabular}{@{}lcccccc@{}}
\toprule
\thead{Variant} & \thead{Avg Turn} & \thead{Guard changes} & \thead{Arbitration} & \thead{Refresh} & \thead{Memory hits} & \thead{Accuracy} \\
\midrule
Full SEAC & 1.00 & 8\% & 3\% & 15\% & 68\% & .713 \\
No clarification & 0.00 & 35\% & 22\% & 0\% & 62\% & .340 \\
No graph memory & 0.03 & 28\% & 18\% & 0\% & 0\% & .707 \\
No uncertainty & 0.00 & 12\% & 8\% & 14\% & 70\% & .710 \\
No experience & 0.38 & 18\% & 10\% & 12\% & 0\% & .297 \\
No retrieval refresh & 0.11 & 10\% & 5\% & 0\% & 66\% & .700 \\
\bottomrule
\end{tabular}
}
\end{table}

\paragraph{Extended visual summaries.}
The main text keeps the formula-level ablation in the paper body because it directly supports the stop-or-clarify boundary. Table~\ref{tab:appendix-equal-budget-visual} expands the equal-budget comparison with a Pareto view; similar turn counts do not imply similar accuracy.

\begin{table}[!htbp]
\centering
\caption{Extended equal-budget visualization. The main text reports the compact numerical comparison; this version adds the Pareto view and budget-effect summary.}
\label{tab:appendix-equal-budget-visual}
\includegraphics[width=0.99\columnwidth]{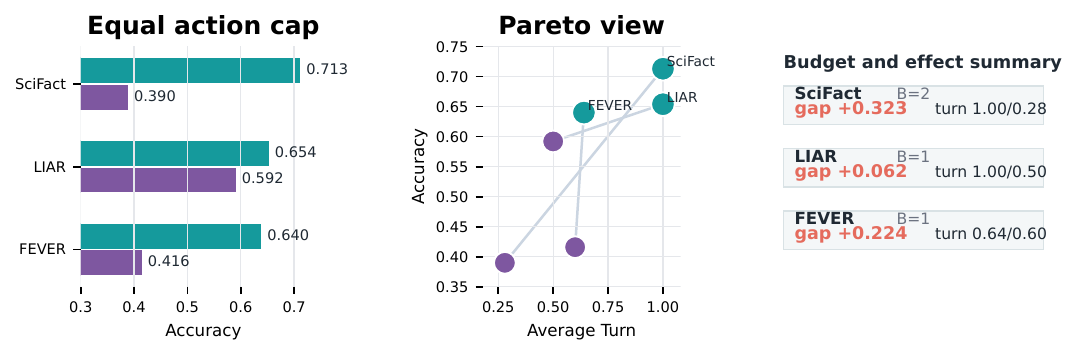}
\end{table}

\paragraph{Offline utility transitions.}
Table~\ref{tab:utility-validity} gives the utility bins summarized in the main text. Higher bins are associated with larger post-hoc accuracy changes after labels are revealed. Table~\ref{tab:utility-transition} gives the label-based transition counts used only for offline analysis. The net SciFact clarification gain is \(52-18=34\) examples, or 11.3\% of the split.

\begin{table}[!htbp]
\centering
\caption{Offline association between label-free trajectory utility and post-hoc label changes for 600 executed clarifications. Labels are used only for analysis, never for decision updates; the bins do not identify a causal clarification effect.}
\label{tab:utility-validity}
\scriptsize
\setlength{\tabcolsep}{5pt}
\renewcommand{\arraystretch}{1.02}
\begin{tabular}{@{}lccc@{}}
\toprule
\thead{\(\delta_i\) bin} & \thead{Mean \(\delta_i\)} & \thead{Acc. gain} & \thead{\(n\)} \\
\midrule
Lowest 20\% & -.12 & -.04 & 120 \\
20--40\% & -.03 & +.01 & 120 \\
40--60\% & +.05 & +.03 & 120 \\
60--80\% & +.15 & +.08 & 120 \\
Highest 20\% & +.32 & +.14 & 120 \\
\bottomrule
\end{tabular}
\end{table}

\begin{table}[!htbp]
\centering
\caption{SciFact clarification transition matrix used only for offline utility validation.}
\label{tab:utility-transition}
\scriptsize
\setlength{\tabcolsep}{6pt}
\renewcommand{\arraystretch}{1.02}
\begin{tabular}{@{}lcc@{}}
\toprule
\thead{Transition} & \thead{Count} & \thead{Percent} \\
\midrule
Correct\(\rightarrow\)Correct & 145 & 48.3\% \\
Wrong\(\rightarrow\)Correct & 52 & 17.3\% \\
Correct\(\rightarrow\)Wrong & 18 & 6.0\% \\
Wrong\(\rightarrow\)Wrong & 85 & 28.3\% \\
\bottomrule
\end{tabular}
\end{table}

\paragraph{Order robustness.}
Table~\ref{tab:order-robustness} reports the effect of random stream order. The variance is visible but moderate: 0.018 on SciFact and 0.015 on FEVER. Table~\ref{tab:learning-curve} reports the typical online progression. Early examples are less stable because the experience store is still sparse.

\begin{table}[!htbp]
\centering
\caption{Online order robustness over five shuffled evaluation orders.}
\label{tab:order-robustness}
\scriptsize
\setlength{\tabcolsep}{6pt}
\renewcommand{\arraystretch}{1.02}
\begin{tabular}{@{}lcccc@{}}
\toprule
\thead{Dataset} & \thead{Acc.} & \thead{Std} & \thead{Turn} & \thead{Std} \\
\midrule
SciFact & .708 & .018 & 1.00 & .00 \\
FEVER & .635 & .015 & 0.64 & .02 \\
\bottomrule
\end{tabular}
\end{table}

\begin{table}[!htbp]
\centering
\caption{SciFact online progression over the evaluation stream.}
\label{tab:learning-curve}
\scriptsize
\setlength{\tabcolsep}{5pt}
\renewcommand{\arraystretch}{1.02}
\begin{tabular}{@{}lccc@{}}
\toprule
\thead{Progress} & \thead{Accuracy} & \thead{Mean \(g_t\)} & \thead{Turn} \\
\midrule
0--50 & .58 & .35 & .95 \\
50--100 & .65 & .48 & .90 \\
100--150 & .70 & .55 & .88 \\
150--200 & .72 & .58 & .85 \\
200--250 & .74 & .60 & .83 \\
250--300 & .75 & .62 & .82 \\
\bottomrule
\end{tabular}
\end{table}

\paragraph{Failure and efficiency diagnostics.}
Table~\ref{tab:failure-decomposition} breaks the remaining errors into wrong stops, unhelpful clarifications, retrieval failures, experience mismatch, and arbitration or guard errors. Table~\ref{tab:runtime-efficiency} reports practical cost: each example uses roughly two LLM calls, fewer than two retrieval calls on average, and a bounded memory footprint.

\begin{table}[!htbp]
\centering
\caption{Quantitative error decomposition.}
\label{tab:failure-decomposition}
\scriptsize
\setlength{\tabcolsep}{4pt}
\renewcommand{\arraystretch}{1.02}
\begin{tabular}{@{}lccc@{}}
\toprule
\thead{Error type} & \thead{SciFact} & \thead{FEVER} & \thead{LIAR} \\
\midrule
Wrong stop & 28\% & 22\% & 35\% \\
Unhelpful clarification & 25\% & 28\% & 20\% \\
Retrieval failure & 20\% & 18\% & 15\% \\
Experience mismatch & 15\% & 18\% & 18\% \\
Arbitration/guard error & 12\% & 14\% & 12\% \\
\bottomrule
\end{tabular}
\end{table}

\begin{table}[!htbp]
\centering
\caption{Runtime and interaction-cost diagnostics.}
\label{tab:runtime-efficiency}
\scriptsize
\setlength{\tabcolsep}{4pt}
\renewcommand{\arraystretch}{1.02}
\resizebox{\columnwidth}{!}{%
\begin{tabular}{@{}lcccccc@{}}
\toprule
\thead{Dataset} & \thead{Latency (s)} & \thead{Retrieval calls} & \thead{LLM calls} & \thead{Tokens} & \thead{Memory size} & \thead{Turn} \\
\midrule
SciFact & 2.8 & 1.8 & 2.0 & 820 & 3.8K & 1.00 \\
FEVER & 2.4 & 1.5 & 1.9 & 750 & 4.1K & 0.64 \\
LIAR & 2.6 & 1.9 & 2.0 & 890 & 4.5K & 1.00 \\
\bottomrule
\end{tabular}
}
\end{table}

The failure categories are decision errors rather than separate model failures. A wrong stop means that the boundary accepted a verifier label before missing evidence was recovered. LIAR has the largest share of this error, which is plausible because short political statements often require background context that is weakly represented in the retrieved corpus. An unhelpful clarification means that the method spent a turn but the refreshed evidence did not improve the final label. FEVER has the clearest version of this problem: additional Wikipedia passages can be topically close while still failing to resolve the exact claim relation. Retrieval failures cover cases in which the generated query is reasonable but the corpus response lacks decisive evidence. Experience mismatch is the complementary case: useful evidence may exist, but the retrieved trajectory neighborhood gives a poor estimate of whether clarification will help.

The efficiency table puts these errors in context. Across datasets, the method uses about two LLM calls per example and fewer than two retrieval calls on average. The memory footprint remains small because records are stored as decision summaries rather than full trajectories. Most added cost comes from one verifier call and one clarification-generation call when the boundary chooses to act. The error profile points to two concrete improvements: better state representations for reducing wrong stops, and better query grounding for reducing unhelpful clarifications.

The appendix is organized as a set of checks on the decision claim. Table~\ref{tab:ablation-audit} explains why coarse switches can be misleading: guard and arbitration paths can alter labels even when the nominal turn count is low. Table~\ref{tab:utility-validity} and Table~\ref{tab:utility-transition} examine the label-free utility proxy after labels are revealed. Their monotonic bins and positive wrong-to-correct balance indicate that the proxy carries useful information, while still falling short of a counterfactual reward. Table~\ref{tab:order-robustness} and Table~\ref{tab:learning-curve} examine stream-order sensitivity. The variance is visible, especially early in the stream, but it does not change the qualitative ranking.

The remaining diagnostics clarify where the boundary succeeds and where it fails. Shuffled-experience and static-gain controls test whether local trajectory matching matters beyond a global prior. The no-cost variant tests whether clarification remains useful when the action is treated as free. The observed pattern supports a limited claim: experience helps when it is matched to the current decision state and when clarification remains costly. The same pattern also identifies the main limitation. The method extrapolates observable outcomes of executed clarifications to nearby states; it does not recover the true counterfactual value of a missed clarification. Future work can replace this estimator with logged bandit feedback or supervised action outcomes when such data are available, while preserving the same stop-or-clarify interface.

For reproducibility, online and analysis-only quantities are kept separate. During inference, the system stores decision states, executed actions, and label-free utility summaries. It never stores the gold label for the current example before the final decision is made. The transition matrix, failure decomposition, and significance tests are computed only after prediction. In this sense, the appendix is not a second set of headline results. It is a record of how the reported decisions were audited.

\end{document}